\documentclass[letterpaper]{article} 
\usepackage[preprint]{aaai2027}
\usepackage[hyphens]{url}  
\usepackage{graphicx} 
\usepackage{natbib}  
\usepackage{caption} 
\usepackage{algorithm}

\usepackage{newfloat}
\usepackage{listings}
\DeclareCaptionStyle{ruled}{labelfont=normalfont,labelsep=colon,strut=off} 
\floatstyle{ruled}
\newfloat{listing}{tb}{lst}{}
\floatname{listing}{Listing}

\usepackage{booktabs,array}

\usepackage{amsmath}
\usepackage{multirow}
\usepackage{amssymb}
\usepackage{mathtools}
\usepackage{tikz}
\usepackage{amsthm}
\usepackage{makecell}
\usepackage{algpseudocode}
\usepackage{enumitem}
\usepackage{microtype}
\usepackage{subcaption}

\newcommand{\dv}[2]{#1\,{\scriptsize$(#2)$}}

\newcommand{\Ex}{\mathbb{E}}

\title{To Go Far, Go Together:\\ Diverse Preferences Induce a Curriculum for Reward Optimization}
\author{
    Taehyung Kim\textsuperscript{\rm 1},
    Jongeun Choi\textsuperscript{\rm 1}\thanks{Corresponding Author.}
}

\affiliations{
    \textsuperscript{\rm 1}Yonsei University, Seoul, South Korea\\
    \{kth8606, jongeunchoi\}@yonsei.ac.kr
}

\begin{document}

\maketitle

\begin{abstract}
Learning a reward model from human feedback and optimizing a policy against it is one approach to aligning AI systems with individual users. From a fairness perspective, existing work improves such alignment by developing data-efficient and accurate reward models that capture minority preferences despite scarce data. We push this line of inquiry one step further and argue that data-efficient and accurate per-user reward models are not sufficient: users whose reward models are difficult to \textit{optimize} at the policy level can become a new underserved group. We start from the observation that one user's reward model can be easy to optimize from the initial policy while another's is not. We argue that, given a sufficiently diverse user population, a curriculum naturally emerges between easy- and hard-to-optimize reward models. Building on this insight, we propose CurriPO, which grows a tree-structured curriculum to accommodate diverse user-specific objectives, covering the population in a single traversal. Specifically, CurriPO automatically constructs a curriculum over diverse user reward models, allowing it to branch from the existing curriculum and reuse reward models previously incorporated into the curriculum. To the best of our knowledge, this is the first work to explicitly exploit multi-user structure to address optimization in AI alignment. Extensive experiments on personalized continuous control in a simulated environment show that CurriPO achieves $1.2$--$2.1\times$ the population satisfaction of the strongest baseline while substantially reducing training time. Additional analysis attributes much of this improvement to the users left underserved by conventional optimization.
\end{abstract}


\section{Introduction}
A widely adopted approach to aligning AI systems with human values is to elicit preference feedback (e.g., pairwise comparisons), fit a reward model, and optimize a policy against it with reinforcement learning (RL) \citep{christiano2017deep}. This approach has since been applied and extended across a range of domains, achieving substantial gains \citep{ouyang2022training, ref10}. As these systems have become widely deployed, there has been a growing demand for alignment at the level of individual users \citep{guan2025survey}. 

\begin{figure}[t]
    \centering
    \includegraphics[width=\linewidth]{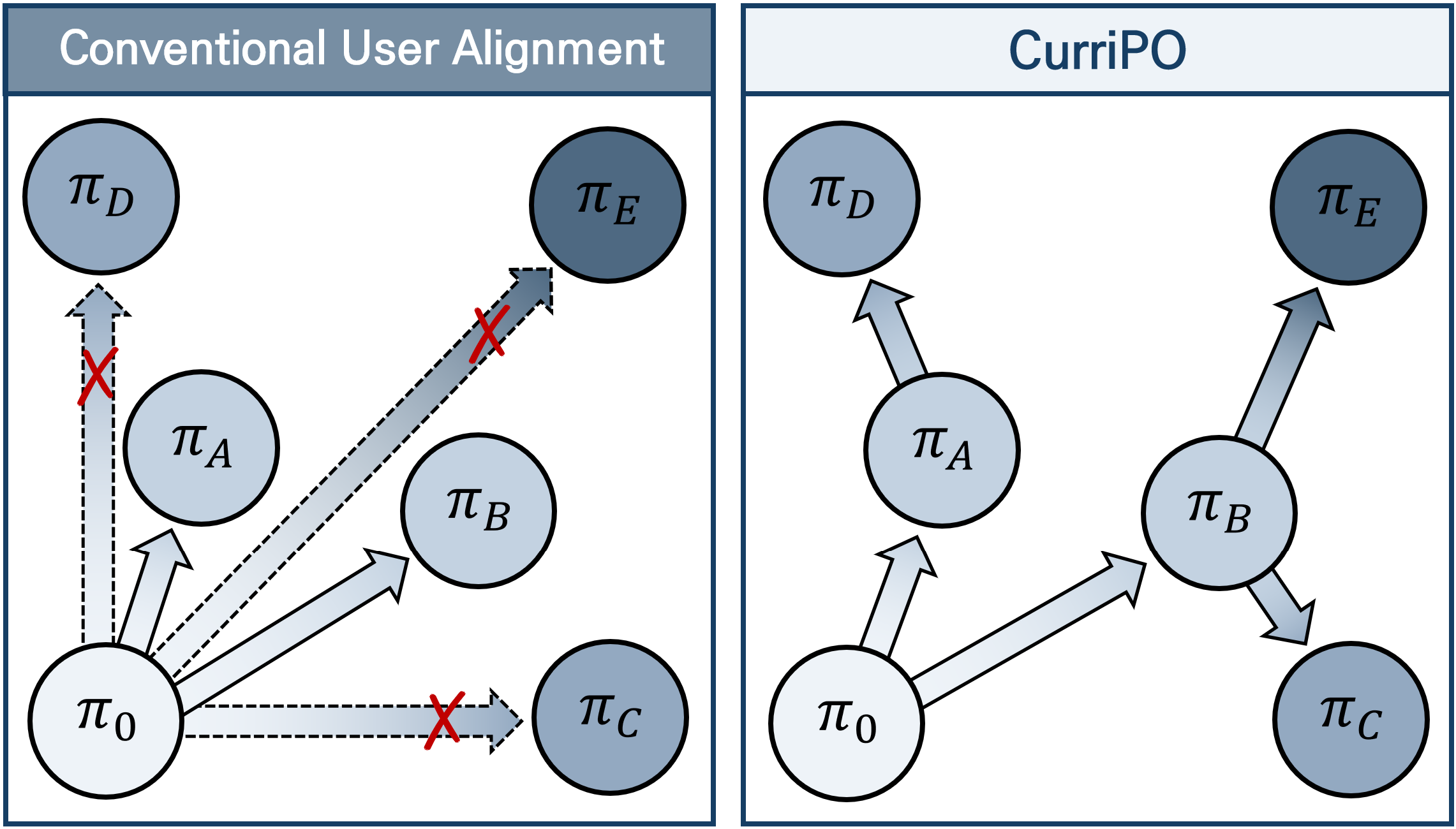}
    \caption{CurriPO aims to construct an effective curriculum across users, enabling better alignment for those who would otherwise remain underserved by conventional methods.}
    \label{fig:example}
\end{figure}

Prior work has developed various reward model-based preference optimization methods to support personalized alignment. These include reward models that capture heterogeneous preferences by modeling distributions over latent utilities \citep{siththaranjan2024distributional}, shared preference-space methods with few-shot onboarding \citep{poddar2024personalizing}, and methods that use limited reward models based on mixtures of preference groups or combinations of shared prototypes \citep{shenfeld2025language, kim2026deployable}. From the perspective of protecting minority preferences, these approaches leverage the structural advantage of multi-user settings to improve personalized reward model accuracy and data efficiency, helping preserve preferences that might otherwise be overlooked.

Although these approaches are effective, we go one step further by highlighting that even when accurate personalized reward models are available for all users, including minorities, subsequent RL optimization can itself create new forms of underrepresentation. When modern methods drive a reward model to high held-out accuracy, it remains unclear whether a policy can be effectively optimized against it. \citet{ng1999policy} characterize a family of reward transformations that preserve the optimal policy while substantially altering the difficulty of learning it. Extending this line of analysis to learned reward functions, \citet{skalse2023invariance} show that accuracy on held-out feedback alone cannot distinguish reward models that are easy to optimize from those that are not. Similarly, \citet{NEURIPS2025_554e056f} show that a reward model can correctly rank every pair while still inducing a flat optimization landscape that makes policy optimization difficult. This optimization issue can be particularly pronounced in per-user alignment, where a separate policy must be optimized for each user under a user-specific reward landscape.

To address the reward optimization problem across a diverse user population, we begin with the observation that reward models learned from different users vary substantially in how easily a policy can be optimized against them. Our key intuition is that users whose reward models are easier to optimize can serve as stepping stones toward users with progressively harder reward models, enabling a curriculum-like optimization procedure. We instantiate this idea as CurriPO, a tree-structured traversal of policy space that treats the diverse reward model population as its curriculum. Starting from an initial policy $\pi_0$, CurriPO repeatedly optimizes a selected user's reward model, stores the resulting policy checkpoint, and selects the next reward model to optimize. All checkpoints generated during traversal are retained in a growing set, which we call the \textit{wake}. At each stage, CurriPO probes the entire wake by running short rollouts and evaluating them under users' reward models, thereby identifying the next reward model–checkpoint pair to optimize. Once the curriculum is complete, each user is ultimately served the checkpoint in the wake that is most preferred by that user's own reward model.

CurriPO incorporates two key mechanisms. First, \textit{branching} allows the next optimization to start from any checkpoint in the wake, rather than only from the most recent one. This prevents optimization from degenerating into a single chain and enables broader coverage of heterogeneous users. Second, \textit{reusing} continues to probe previously optimized reward models, allowing them to re-enter the curriculum when probing identifies a later checkpoint as a more promising initialization. Thus, a reward model previously difficult to optimize can be reused from a better initialization. Through these mechanisms, CurriPO turns the multiplicity of users---often viewed as an obstacle to per-user alignment---into a sequence of learning stages. By explicitly leveraging this multi-user structure, CurriPO reduces the risk that users associated with harder-to-optimize reward models are systematically underserved during the alignment process.

We consider a setting in which CurriPO optimizes a policy using pairwise preferences collected from users. In our evaluation, we focus on personalizing relatively simple robots that are likely to be deployed around humans in the near term, instead of highly complex systems such as humanoid robots. We study personalization on continuous-control tasks in MuJoCo \citep{todorov2012mujoco}. We show that CurriPO substantially improves population satisfaction while reducing training time relative to both conventional reward model optimization and direct preference optimization baselines. Ablation studies further validate the contributions of its key components. Additional analysis demonstrates that CurriPO provides particularly large gains for users who are poorly served by existing optimization methods. 

Our key contributions are summarized as follows:
\begin{itemize}
    \item We propose CurriPO, a novel method for improving optimization in per-user alignment in multi-user settings, which treats the diverse user population as a curriculum to identify favorable optimization trajectories for users whose reward models are difficult to optimize.
    \item CurriPO admits branching and reusing when ordering the curriculum, yielding a tree-structured curriculum that serves all users in a single traversal and thus achieves more efficient and accurate user alignment.
    \item Extensive experiments in continuous control show that CurriPO improves population satisfaction over conventional optimization baselines, with the largest gains observed for users whose reward models are difficult to optimize against using conventional methods.
\end{itemize}

\section{Related Work}
\textbf{Preference-Based Policy Learning and Alignment.} 
Preference-based RL replaces manually specified rewards with human preferences \citep{christiano2017deep}. This paradigm has been steadily refined across diverse domains of continuous control \citep{lee2021pebble, ref21}. As preference feedback increasingly comes from heterogeneous populations, recent work has sought to account for individual differences. \citet{poddar2024personalizing} personalize robot control policies by inferring a per-user latent preference within a shared latent space, and, in a similar spirit, \citet{kalrap3l} decompose the preferences of multiple users into a common preference basis. \citet{kim2026deployable} group users with similar preferences and learn a reward model for each group, aiming to obtain robot policies that serve the population well. While these methods assume a multi-user setting and devote their efforts to constructing better reward models, whether multiple users can also be leveraged to improve the policy optimization stage itself has rarely been explored. To the best of our knowledge, CurriPO is the first to address this gap, using policies obtained by optimizing one user's reward as stepping stones that make other users' reward models easier to optimize.

A parallel line of work bypasses explicit reward modeling and optimizes policies directly from preferences. IPL \citep{hejna2023inverse} learns a preference-consistent Q-function via an implicit reward, while CPL \citep{hejna2024contrastive} directly optimizes the policy with a contrastive objective. However, removing the explicit reward model does not eliminate optimization difficulties: direct preference optimization remains constrained by the coverage of the preference data and by the policy distribution from which optimization proceeds \citep{razin2025unintentional, guo2025role}. Discarding the reward model therefore does not resolve the problem CurriPO targets.

\smallskip
\noindent \textbf{Curriculum Learning.} 
Curriculum learning organizes training examples or tasks to match the learner's evolving capabilities \citep{bengio2009curriculum, kumar2010self}, with automatic variants adapting the curriculum using signals such as learning progress \citep{graves2017automated, matiisen2019teacher}. In RL, self-paced RL adapts task distributions to the agent's competence \citep{klink2020self}, while PLR \citep{jiang2021prioritized} prioritizes levels with high learning potential. Unsupervised Environment Design further generates or curates tasks near the learning frontier; examples include PAIRED \citep{dennis2020emergent}, replay-guided environment design \citep{jiang2021replay}, and ACCEL \citep{parker2022evolving}. Recent methods improve task generation and selection through learned task representations \citep{azad2023clutr} or cross-task learnability \citep{cho2026traced}. Related open-ended methods such as POET transfer policies across challenges, using intermediate solutions as stepping stones toward otherwise difficult behaviors \citep{wang2019paired, wang2020enhanced}.

CurriPO differs in two key respects. First, it requires no separate process for designing or generating curriculum tasks: the heterogeneous reward models already present in the user population form the task space, allowing population diversity to be directly exploited as a structural resource for optimization. Second, CurriPO is not a curriculum toward a single target. Through branching and reusing, it expands toward multiple user-specific objectives, effectively solving many user alignment problems within a shared traversal. 

\section{Method}
\subsection{Problem formulation}
We aim to serve a diverse population with user-aligned policies, derived from $\pi_0$, using only pairwise user feedback collected over a behavior pool $\mathcal{D}$. Let
$\mathcal{M}=(\mathcal{S},\mathcal{A},P,\rho_0,H)$ be a reward-free episodic decision process, where a policy $\pi(a\mid s)$ specifies the probability of choosing action $a$ in state $s$, and executing the policy yields a trajectory $\tau=(s_0,a_0,\dots,s_{H-1},a_{H-1})\in\mathcal{T}$. Let
$\mathcal{U}=\{1,\dots,N\}$ be the population of users. We assume each user $u$ carries a latent utility $g_u$ capturing how satisfied $u$ actually is with a given behavior. From the feedback $\mathcal{F}_u$ that $u$ provides on trajectory segments $\xi=((s_t,a_t))_{t=0}^{L-1}$, we fit a per-user reward model $f_u:\mathcal{S}\times\mathcal{A}\to\mathbb{R}$, whose segment score $\bar f_u(\xi)=\frac{1}{L}\sum_t f_u(s_t,a_t)$ is a learned proxy for $g_u$. We call $\{f_u\}_{u\in\mathcal{U}}$ the reward model bank. Given the bank, we aim to provide each user $u$ with a personalized policy $\pi_u$, improving overall population satisfaction $\frac{1}{N}\sum_{u}\Ex_{\xi\sim\pi_u}[g_u(\xi)]$ while better serving users who would otherwise remain underserved by conventional policy optimization. 

\subsection{Reward modeling from pairwise preferences}
We instantiate the bank following the standard preference-based pipeline \citep{christiano2017deep}. Each element of $\mathcal{F}_u$ is a triple $(\xi^0,\xi^1,y)$ with $y\in\{0,1\}$, where $y=1$ denotes $\xi^1\succ\xi^0$ and $y=0$ denotes $\xi^0\succ\xi^1$. To fit $f_u$, most prior work \citep{christiano2017deep, lee2021pebble} defines a preference predictor following the Bradley--Terry model \citep{bradley1952rank}; we adopt the predictor per user, expressed as $P_{f_u}\!\left[\xi^1\succ\xi^0\right]=\mathrm{logistic}\big(\bar f_u(\xi^1)-\bar f_u(\xi^0)\big)$. Each $f_u$ is fit by minimizing the binary cross-entropy below, and repeating this for every $u\in\mathcal{U}$ yields the reward model bank:
\begin{equation}
\begin{aligned}
\mathcal{L}^{\mathrm{CE}}(f_u)=-\underset{(\xi^0,\xi^1,y)\sim\mathcal{F}_u}{\Ex}\Big[
&(1-y)\log P_{f_u}\!\left[\xi^0\succ\xi^1\right]\\
&+\,y\log P_{f_u}\!\left[\xi^1\succ\xi^0\right]\Big].
\end{aligned}
\label{eq:bcce}
\end{equation}

\begin{figure*}[t]
    \centering
    \includegraphics[width=0.9\linewidth]{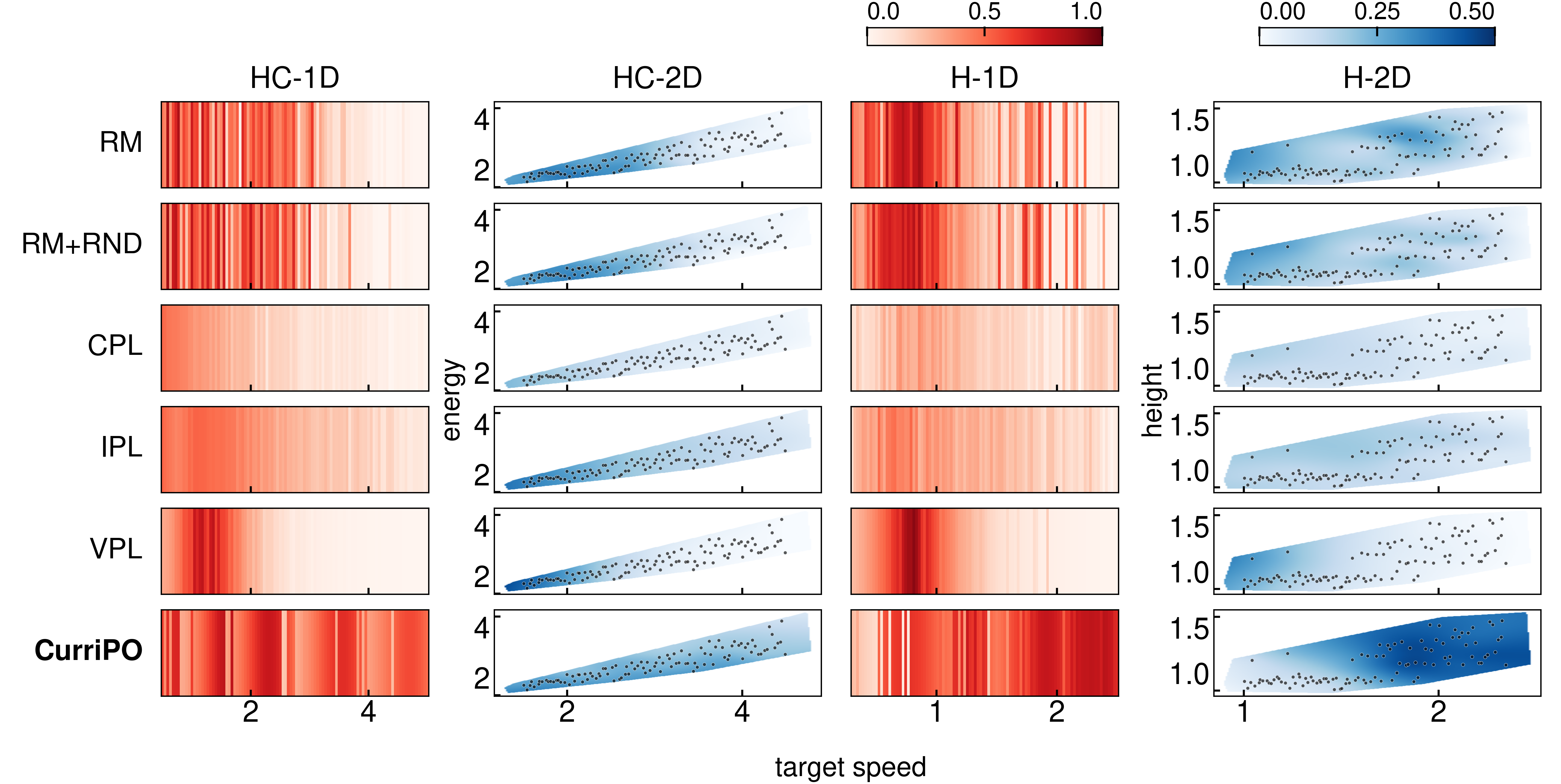}
    \caption{\textbf{User satisfaction visualization across different optimization methods with $N=100$ users under oracle feedback.} User satisfaction landscapes are shown for HalfCheetah and Hopper under 1-D and 2-D preference settings, with target speed as the shared preference dimension and energy or height as the second dimension. Darker colors indicate higher user satisfaction.}
    \label{figure_2}
\end{figure*}

\subsection{CurriPO: Probe-Driven Reward Traversal}
\label{sec:curripo}
CurriPO exploits the reward model bank as an automatically discovered curriculum. Starting from $\pi_0$, CurriPO traverses the policy space by iteratively optimizing a selected user's reward model, retaining the resulting policy checkpoint, and selecting the reward model for the next stage. After $k$ stages, these checkpoints form the \emph{wake} $\mathcal{W}_k=\{\pi_0,\ldots,\pi_k\}$. At stage $k$, CurriPO selects $(w_k,u_k)$, where $\pi_{w_k}\in\mathcal{W}_k$ is the checkpoint used to warm-start optimization of the selected reward model $f_{u_k}$.

\paragraph{Probing the wake.}
CurriPO evaluates each checkpoint using short probe rollouts, scored by every user reward model. For each checkpoint--user pair, we estimate the \emph{competence}
\begin{equation}
  \operatorname{comp}(\pi,u)
  = \widehat{\mathbb{E}}_{\xi\sim\pi}
    \!\left[\mathcal{N}_u\!\left(\bar f_u(\xi)\right)\right],
  \label{eq:comp}
\end{equation}

where $\mathcal{N}_u(x) = P_{\xi \sim \mathcal{D}}[\bar{f}_u(\xi) \le x]$
is the empirical cumulative distribution function over the pool $\mathcal{D}$, so competence is the pool percentile $\pi$ reaches under $u$. Below we write $\operatorname{comp}_w(u)$ for the competence of checkpoint $\pi_w$ evaluated under user $u$'s reward model. Aggregating competence over the wake gives the coverage
\begin{equation}
  C_k(u) = \max_{w\,:\,\pi_w\in\mathcal{W}_k}\operatorname{comp}_w(u),
  \label{eq:coverage}
\end{equation}
so that $C_k(u)$ measures how well user $u$ is served by the best checkpoint currently in the wake.
 
\paragraph{Selecting the next traversal step.}
Having probed the wake, CurriPO has an estimate of how well each user is currently served, summarized by the coverage $C_k(u)$. It then uses this coverage information to decide where the traversal should expand next. CurriPO follows a curriculum principle of expanding progressively toward reward models that are sufficiently accessible from the current wake \citep{florensa2018automatic}. To this end, it first restricts attention to the admissible users, namely those some checkpoint in the wake already serves above a minimum threshold $c_{\mathrm{lo}}$:
\begin{equation}
\mathcal{U}_k=\big\{u:\ C_k(u)\ge c_{\mathrm{lo}}\big\}.
\end{equation}
Among these, CurriPO targets the least-covered user and warm-starts from the checkpoint that currently serves that user best:
\begin{equation}
u_k=\arg\min_{u\in\mathcal{U}_k}C_k(u),
\quad
w_k=\arg\max_{w\,:\,\pi_w\in\mathcal{W}_k}\operatorname{comp}_w(u_k),
\label{eq:pick}
\end{equation}
with ties broken toward the most recently added checkpoint. 

This design enables CurriPO to select the starting checkpoint afresh at every stage, without requiring every reward optimization stage to start from the immediately preceding checkpoint. As a result, the traversal can branch into a tree instead of being a single chain. This is a particularly effective design choice for our setting, where the goal is not to reach a single difficult objective, but to efficiently cover a diverse population of users with different hard-to-optimize objectives.

\begin{figure}[t]
    \centering
    \includegraphics[width=\linewidth]{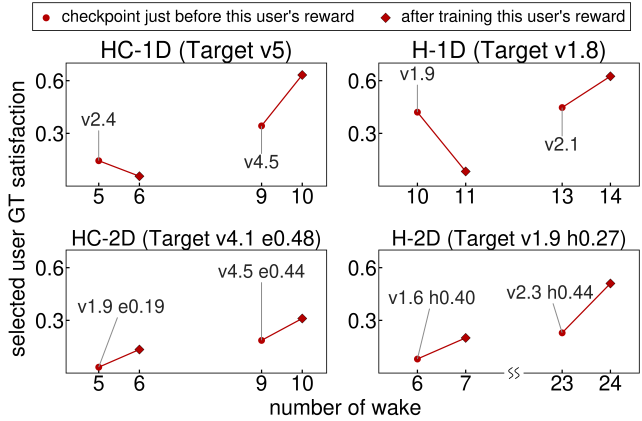}
    \caption{\textbf{Case studies of rescuing users from local optima.} The x-axis denotes the wake number, which increases as the CurriPO tree progressively branches over training. The y-axis denotes the selected user’s ground-truth satisfaction. Circles indicate the checkpoint used to warm-start training for the selected user, with the preceding user’s target additionally provided for reference; v, e, and h denote target speed, target energy, and target height, respectively. Diamonds indicate the checkpoint obtained after training the selected user’s reward model starting from the corresponding warm-start policy.}
    \label{figure_4}
\end{figure}

Additionally, motivated by the observation that optimizing the same reward model from different initializations can induce substantially different optimization dynamics \citep{NEURIPS2025_554e056f}, which we also empirically observe in Figure~\ref{figure_4}, we allow each reward model that has already been selected and used for optimization to be reused up to $q$ times. CurriPO gives a previously selected reward model another opportunity for optimization whenever a newly added checkpoint in the wake achieves a better probe score than the checkpoint from which that reward model was previously optimized. This mechanism is particularly well suited to our setting, where stepping-stone checkpoints remain relevant even after the curriculum is complete because they are retained as candidate policies for final user assignment. Reusing reward models can therefore improve the quality of these stepping stones while simultaneously inducing a finer-grained curriculum as the wake expands.

\paragraph{Local stage optimization.}
Once $(w_k,u_k)$ is selected, CurriPO initializes the policy from $\pi_{w_k}$ and performs a local optimization stage:
\begin{equation}
  \pi_{k+1} \approx \arg\max_{\pi}\;
  \mathbb{E}_{\tau\sim\pi} \sum_t
  [\, \hat r_{u_k} - \beta\, D_{\mathrm{KL}}(\pi \,\|\, \pi_{w_k}) \,].
  \label{eq:stage}
\end{equation}
Here, $\pi_{w_k}$ is kept fixed as the reference policy throughout the stage. 
To make the shared regularization coefficient $\beta$ comparable across users, 
we standardize each reward model over the pool $\mathcal{D}$: 

\begin{equation}
  \hat r_u(s,a)
  =
  \frac{f_u(s,a)-\mu_u^{\mathcal D}}
       {\sigma_u^{\mathcal D}}.
  \label{eq:zscore}
\end{equation}
The KL penalty preserves useful behaviors from the selected checkpoint while preventing each optimization stage from moving too far from its initialization. Re-anchoring the constraint at successive checkpoints allows CurriPO to progressively traverse policy space through local transitions.

\paragraph{Serving from the wake.}
The traversal terminates when $\min_u C_k(u)\ge c_{\mathrm{stop}}$ or when the interaction budget $B$ is exhausted. The final output is the entire wake. Each user $u$ is assigned the checkpoint preferred by that user's own reward model,

\begin{equation}
\pi_u \;=\; \arg\max_{\pi\in\mathcal{W}}\ \operatorname{comp}(\pi,u).
\label{eq:serve}
\end{equation}
 
\begin{table*}[t]
\centering\small

\setlength{\tabcolsep}{4pt}
\renewcommand{\arraystretch}{0.9}
\setlength{\aboverulesep}{1pt}
\setlength{\belowrulesep}{1pt}

\begin{tabular}{lccccccccc}
\toprule
\multirow{2}{*}{Method} & \multicolumn{4}{c}{Oracle} & \multicolumn{4}{c}{B-Pref noise} & \multirow{2}{*}{Time (h)} \\
\cmidrule(lr){2-5} \cmidrule(lr){6-9}
& HC-1D & HC-2D & H-1D & H-2D & HC-1D & HC-2D & H-1D & H-2D & \\
\midrule
RM         & 0.341{\scriptsize$\pm$0.004} & 0.215{\scriptsize$\pm$0.006} & 0.250{\scriptsize$\pm$0.027} & 0.175{\scriptsize$\pm$0.001} & 0.311{\scriptsize$\pm$0.012} & 0.220{\scriptsize$\pm$0.002} & 0.309{\scriptsize$\pm$0.012} & 0.163{\scriptsize$\pm$0.011} & 2.6 \\
RM+RND     & 0.388{\scriptsize$\pm$0.000} & 0.239{\scriptsize$\pm$0.013} & 0.260{\scriptsize$\pm$0.003} & 0.157{\scriptsize$\pm$0.001} & 0.320{\scriptsize$\pm$0.003} & 0.168{\scriptsize$\pm$0.002} & 0.312{\scriptsize$\pm$0.027} & 0.156{\scriptsize$\pm$0.009} & 3.4 \\
CPL            & 0.188{\scriptsize$\pm$0.010} & 0.091{\scriptsize$\pm$0.008} & 0.188{\scriptsize$\pm$0.020} & 0.076{\scriptsize$\pm$0.008} & 0.194{\scriptsize$\pm$0.003} & 0.096{\scriptsize$\pm$0.006} & 0.180{\scriptsize$\pm$0.005} & 0.085{\scriptsize$\pm$0.005} & 5.6 \\
IPL            & 0.295{\scriptsize$\pm$0.007} & 0.193{\scriptsize$\pm$0.007} & 0.228{\scriptsize$\pm$0.007} & 0.125{\scriptsize$\pm$0.008} & 0.304{\scriptsize$\pm$0.001} & 0.193{\scriptsize$\pm$0.008} & 0.253{\scriptsize$\pm$0.008} & 0.128{\scriptsize$\pm$0.009} & 5.7 \\
VPL            & 0.209{\scriptsize$\pm$0.009} & 0.087{\scriptsize$\pm$0.013} & 0.266{\scriptsize$\pm$0.001} & 0.099{\scriptsize$\pm$0.005} & 0.192{\scriptsize$\pm$0.021} & 0.114{\scriptsize$\pm$0.023} & 0.264{\scriptsize$\pm$0.016} & 0.098{\scriptsize$\pm$0.006} & 3.5 \\
CurriPO        & \textbf{0.514}{\scriptsize$\pm$0.016} & \textbf{0.318}{\scriptsize$\pm$0.040} & \textbf{0.355}{\scriptsize$\pm$0.059} & \textbf{0.328}{\scriptsize$\pm$0.006} & \textbf{0.386}{\scriptsize$\pm$0.034} & \textbf{0.344}{\scriptsize$\pm$0.009} & \textbf{0.371}{\scriptsize$\pm$0.021} & \textbf{0.340}{\scriptsize$\pm$0.038} & \textbf{1.1} \\
\bottomrule
\end{tabular}
\caption{Satisfaction and training time for $N=12$ users under oracle and B-Pref noisy feedback across three seeds.}
\label{table_1}
\end{table*}

\begin{table*}[t]
\centering\small
\setlength{\tabcolsep}{4pt}
\renewcommand{\arraystretch}{0.9}
\setlength{\aboverulesep}{1pt}
\setlength{\belowrulesep}{1pt}
\begin{tabular}{lcccccccc@{\hspace{12pt}}c}
\toprule
\multirow{2}{*}{Method} & \multicolumn{4}{c}{Oracle} & \multicolumn{4}{c}{B-Pref noise} & \multirow{2}{*}{Time (h)} \\
\cmidrule(lr){2-5} \cmidrule(lr){6-9}
& HC-1D & HC-2D & H-1D & H-2D & HC-1D & HC-2D & H-1D & H-2D & \\
\midrule
RM         & 0.352{\scriptsize$\pm$0.000} & 0.227{\scriptsize$\pm$0.001} & 0.309{\scriptsize$\pm$0.002} & 0.142{\scriptsize$\pm$0.003} & 0.318{\scriptsize$\pm$0.003} & 0.183{\scriptsize$\pm$0.001} & 0.244{\scriptsize$\pm$0.007} & 0.179{\scriptsize$\pm$0.004} & 22 \\
RM+RND     & 0.344{\scriptsize$\pm$0.005} & 0.221{\scriptsize$\pm$0.000} & 0.318{\scriptsize$\pm$0.008} & 0.145{\scriptsize$\pm$0.005} & 0.316{\scriptsize$\pm$0.005} & 0.171{\scriptsize$\pm$0.004} & 0.267{\scriptsize$\pm$0.003} & 0.152{\scriptsize$\pm$0.004} & 28 \\
CPL            & 0.173{\scriptsize$\pm$0.015} & 0.089{\scriptsize$\pm$0.011} & 0.191{\scriptsize$\pm$0.017} & 0.078{\scriptsize$\pm$0.007} & 0.179{\scriptsize$\pm$0.002} & 0.093{\scriptsize$\pm$0.001} & 0.194{\scriptsize$\pm$0.006} & 0.087{\scriptsize$\pm$0.006} & 47 \\
IPL            & 0.299{\scriptsize$\pm$0.008} & 0.195{\scriptsize$\pm$0.005} & 0.240{\scriptsize$\pm$0.006} & 0.128{\scriptsize$\pm$0.006} & 0.296{\scriptsize$\pm$0.007} & 0.191{\scriptsize$\pm$0.004} & 0.261{\scriptsize$\pm$0.018} & 0.129{\scriptsize$\pm$0.002} & 48 \\
VPL            & 0.208{\scriptsize$\pm$0.020} & 0.107{\scriptsize$\pm$0.031} & 0.275{\scriptsize$\pm$0.002} & 0.094{\scriptsize$\pm$0.003} & 0.215{\scriptsize$\pm$0.004} & 0.125{\scriptsize$\pm$0.030} & 0.275{\scriptsize$\pm$0.006} & 0.093{\scriptsize$\pm$0.003} & 28 \\
CurriPO & \textbf{0.512}{\scriptsize$\pm$0.030} & \textbf{0.319}{\scriptsize$\pm$0.019} & \textbf{0.415}{\scriptsize$\pm$0.051} & \textbf{0.280}{\scriptsize$\pm$0.008} & \textbf{0.522}{\scriptsize$\pm$0.018} & \textbf{0.253}{\scriptsize$\pm$0.026} & \textbf{0.350}{\scriptsize$\pm$0.018} & \textbf{0.269}{\scriptsize$\pm$0.012} & \textbf{1.6} \\
\bottomrule
\end{tabular}
\caption{Satisfaction and training time for $N=100$ users under oracle and B-Pref noisy feedback across three seeds.}
\label{table_2}
\end{table*}

\begin{figure*}[t]
    \centering
    \includegraphics[width=0.9\linewidth]{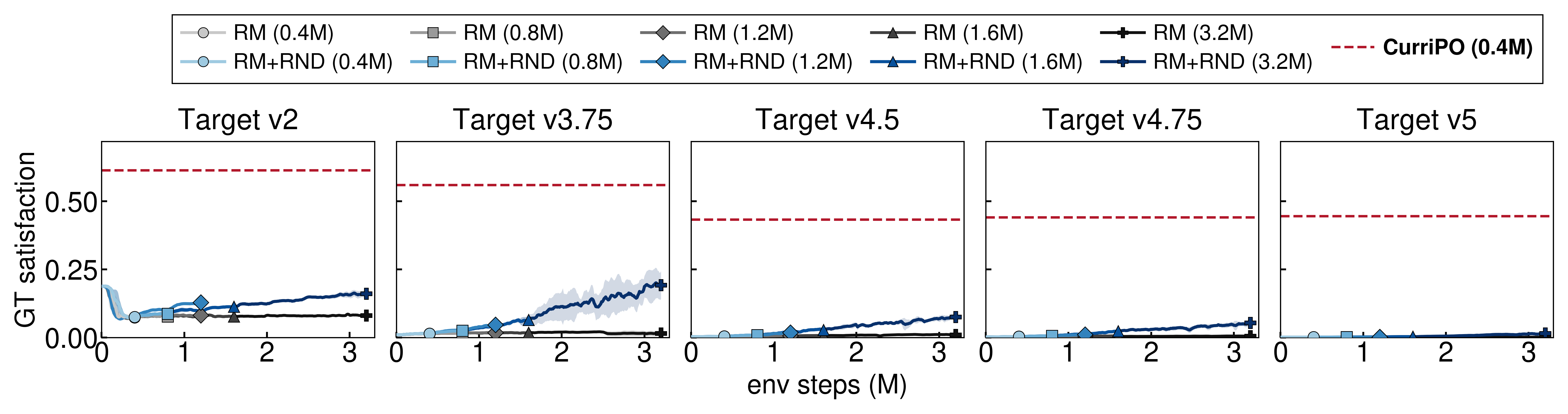}
    \caption{\textbf{Comparison of learning curves with stronger PPO baselines in HC-1D with 12 users.} Ground-truth goal satisfaction over three seeds is shown across different target preferences, comparing RM and RM+RND with increased environment steps.}
    \label{figure_3}
\end{figure*}

\section{Experiments}


\subsection{Experimental Settings}
\textbf{Environment.} Our environment design follows the objective decomposition in MO-Gymnasium \citep{felten_toolkit_2023}, which has been used to study trade-offs among multiple control objectives, and is conceptually aligned with \citet{hwang2024promptable}, who model user preferences through combinations of such objectives. We evaluate the styles preferred by multiple scripted users across four environments: (i) \textbf{HC-1D}, with HalfCheetah as the agent, where user preferences are defined over its velocity; (ii) \textbf{HC-2D}, also using HalfCheetah, where preferences depend jointly on velocity and energy consumption; (iii) \textbf{H-1D}, with Hopper as the agent, where preferences are defined over its velocity; and (iv) \textbf{H-2D}, also using Hopper, where preferences depend jointly on velocity and vertical position (z-height).

\smallskip
\noindent \textbf{Datasets.} We construct a custom dataset $\mathcal{D}$ covering a diverse range of locomotion styles, as existing public offline RL datasets do not provide such coverage. For each environment, we train 10 PPO \citep{schulman2017proximal} policies with different random seeds for 600K steps using velocity-based reward functions with varying target values. We collect trajectories from nine checkpoints throughout training for each policy. The trajectories are then partitioned into non-overlapping 25-step segments and augmented with segments generated by a random policy. Finally, we construct a pairwise dataset by sampling segment pairs from the resulting pool.

\noindent \textbf{User Sampling.}
Following common practice in prior work on personalized preference learning, we use scripted users to evaluate personalization across heterogeneous preferences \citep{poddar2024personalizing, zollo2025personalllm, bahlous2024pareto, kim2026deployable}. In the 1-D settings (HC-1D and H-1D), users are sampled uniformly at random from a predefined speed range. In the 2-D settings (HC-2D and H-2D), users are sampled only from the feasible region. For example, in HC-2D, mismatched speed--energy combinations, such as high speed with low energy or low speed with high energy, are difficult to obtain under our data-collection procedure and are therefore excluded by construction. We conduct experiments with two population sizes, $N=12$ and $N=100$, for each environment. Figures in Appendix~\ref{app:satisfaction-landscape} and Figure~\ref{figure_2} visualize the preference ranges and user populations in the 1-D and 2-D settings for $N=12$ and $N=100$, respectively. To mitigate the limitations of scripted teachers, we further evaluate robustness to imperfect preference feedback by introducing noisy preference labels following the preference-noise setting of B-Pref \citep{lee2021bpref}, yielding reward model geometries that deviate from the oracle geometry \citep{ref23, ref21, ref22}. The preference-noise parameters are detailed in Appendix~\ref{app:bpref}.

\smallskip
\noindent\textbf{Evaluation Metrics.}
We evaluate each personalized policy using the user's ground-truth utility, which is not directly observed during training. For user $u$, the per-step utility is defined as a band-shaped kernel over the behavioral feature vector $d_t$ (e.g., velocity and energy consumption) generated by the policy assigned to the user,
\begin{equation}
  g_u(d_t) \;=\; \exp\!\left(
    -\,\frac{\bigl\lVert (d_t - c_u) \oslash \eta_u \bigr\rVert_2^{2}}{b_u}
  \right),
  \label{eq:per-step-utility}
\end{equation}
where $c_u$, $\eta_u$, and $b_u$ are the user's preferred target, per-dimension tolerance, and bandwidth, and $\oslash$ denotes elementwise division. The policy-level utility $G_u(\pi)$ is the temporal average of $g_u(d_t)$ over fresh stochastic rollouts from $\pi$. Rather than using raw distance in the behavioral descriptor space, we normalize deviations by user-specific tolerances to make heterogeneous preference dimensions comparable. The exponential kernel then maps this normalized distance to a bounded satisfaction score, smoothly capturing how utility decreases as behavior moves away from the user's preferred region \citep{poddar2024personalizing}. Our primary metric is population satisfaction, i.e., the utilitarian population satisfaction $W_{\mathrm{util}} = \frac{1}{N}\sum_{u=1}^{N} G_u(\pi_u)$, where $\pi_u$ is the policy served to user $u$, so that all users contribute equally. Each $G_u$ is estimated from $10$ fresh evaluation episodes. We also report the egalitarian population satisfaction $\min_u G_u(\pi_u)$ and the Nash population satisfaction $\bigl(\prod_u G_u(\pi_u)\bigr)^{1/N}$ in the Appendix~\ref{app:welfare}.

\noindent \textbf{Implementation Details.}
CurriPO is implemented in Python 3.11.15 with PyTorch 2.13.0 and MuJoCo 3.10.0 on Rocky Linux 9.4. All continuous-control training ran on Intel Xeon Gold 5320 CPUs, with each job allocated two CPU cores and 6\,GB RAM. Reward model optimization methods share one PPO implementation: actor and critic MLPs of width 64, 8 parallel environments, 256-step rollouts, 10 epochs, Adam at $3{\times}10^{-4}$, $\gamma{=}0.99$, GAE $\lambda{=}0.95$ and clip $0.2$. For CurriPO, we set $c_{\mathrm{lo}}{=}0.6$, $\beta{=}0.01$, $q{=}2$, and $c_{\mathrm{stop}}{=}0.9$. Each traversal stage is allocated a PPO budget of 400K environment steps. We set a safety cap of $B{=}9.6$M environment steps, but this limit was rarely reached. For probing, we run six 1,000-step episodes from each checkpoint. $\pi_0$ is randomly initialized and held fixed across all methods within a run. As assumed throughout the paper, each reward model is trained until it reaches at least 90\% held-out preference accuracy against oracle labels. All baselines that rely on an explicit reward model use the same reward model bank as CurriPO. 

\subsection{Comparison with State-of-the-Art Methods}
\noindent\textbf{Baselines.} We compare against two families of methods. \emph{Reward model optimization}: \textbf{RM} \citep{christiano2017deep} optimizes each user's reward model directly with PPO, i.e.\ the conventional per-user optimization pipeline; \textbf{RM+RND} adds a random network distillation (RND) bonus \citep{burda2018exploration} to \textbf{RM}, to examine whether insufficient exploration accounts for its poor performance; and \textbf{VPL} \citep{poddar2024personalizing} infers a per-user latent preference and conditions a shared reward model on it. \emph{Direct preference optimization}: \textbf{CPL} \citep{hejna2024contrastive} directly optimizes the policy using a contrastive preference objective, whereas \textbf{IPL} \citep{hejna2023inverse} learns a preference-consistent Q-function through an implicit reward induced by the inverse Bellman operator. Both methods learn policies from preferences without fitting an explicit reward model. Further implementation details for baselines are provided in Appendix~\ref{app:baselines}.

\noindent\textbf{Experiment Results.} 
Table~\ref{table_1} reports population satisfaction for $N{=}12$. Under oracle feedback, CurriPO achieves the highest satisfaction in every environment, with particularly large margins in the more challenging 2-D settings, while no single baseline consistently dominates. These gains persist under B-Pref noise: although noisy feedback affects absolute performance, CurriPO remains the best-performing method across all four environments, indicating that its advantage holds even under the altered reward model geometry induced by imperfect preference feedback.

\begin{table*}
\setlength{\abovecaptionskip}{6pt}   
\setlength{\belowcaptionskip}{0pt}

\renewcommand{\arraystretch}{0.9}
\setlength{\aboverulesep}{1pt}
\setlength{\belowrulesep}{1pt}
\begin{tabular}{lcccccccc}
\toprule
\multirow{2}{*}{Method} & \multicolumn{4}{c}{Oracle} & \multicolumn{4}{c}{B-Pref noise} \\
\cmidrule(lr){2-5} \cmidrule(lr){6-9}
& HC-1D & HC-2D & H-1D & H-2D & HC-1D & HC-2D & H-1D & H-2D \\
\midrule
Chain
& 0.389{\scriptsize$\pm$0.037}
& 0.249{\scriptsize$\pm$0.030}
& 0.324{\scriptsize$\pm$0.013}
& 0.157{\scriptsize$\pm$0.012}
& 0.301{\scriptsize$\pm$0.024}
& 0.223{\scriptsize$\pm$0.008}
& 0.307{\scriptsize$\pm$0.006}
& 0.161{\scriptsize$\pm$0.009} \\

No reuse
& \textbf{0.514}{\scriptsize$\pm$0.016}
& \textbf{0.318}{\scriptsize$\pm$0.040}
& 0.352{\scriptsize$\pm$0.190}
& 0.304{\scriptsize$\pm$0.015}
& 0.352{\scriptsize$\pm$0.082}
& \textbf{0.344}{\scriptsize$\pm$0.009}
& 0.370{\scriptsize$\pm$0.081}
& 0.319{\scriptsize$\pm$0.029} \\

 No KL
& 0.244{\scriptsize$\pm$0.139}
& 0.253{\scriptsize$\pm$0.132}
& \textbf{0.362}{\scriptsize$\pm$0.138}
& 0.320{\scriptsize$\pm$0.016}
& 0.286{\scriptsize$\pm$0.105}
& 0.293{\scriptsize$\pm$0.059}
& 0.358{\scriptsize$\pm$0.102}
& \textbf{0.360}{\scriptsize$\pm$0.052} \\

CurriPO
& \textbf{0.514}{\scriptsize$\pm$0.016}
& \textbf{0.318}{\scriptsize$\pm$0.040}
& 0.355{\scriptsize$\pm$0.059}
& \textbf{0.328}{\scriptsize$\pm$0.006}
& \textbf{0.386}{\scriptsize$\pm$0.034}
& \textbf{0.344}{\scriptsize$\pm$0.009}
& \textbf{0.371}{\scriptsize$\pm$0.021}
& 0.340{\scriptsize$\pm$0.038} \\
\bottomrule
\end{tabular}
\caption{Ablation on branching and reusing for $N=12$ across three seeds.}
\label{table_7}
\end{table*}

\begin{table*}
\setlength{\abovecaptionskip}{6pt}   
\setlength{\belowcaptionskip}{0pt}
\renewcommand{\arraystretch}{0.9}
\setlength{\aboverulesep}{1pt}
\setlength{\belowrulesep}{1pt}
\begin{tabular}{lcccccccc}
\toprule
\multirow{2}{*}{Method} & \multicolumn{4}{c}{Oracle} & \multicolumn{4}{c}{B-Pref noise} \\
\cmidrule(lr){2-5} \cmidrule(lr){6-9}
& HC-1D & HC-2D & H-1D & H-2D & HC-1D & HC-2D & H-1D & H-2D \\
\midrule
Chain
& 0.311{\scriptsize$\pm$0.012}
& 0.222{\scriptsize$\pm$0.008}
& 0.081{\scriptsize$\pm$0.001}
& 0.169{\scriptsize$\pm$0.018}
& 0.280{\scriptsize$\pm$0.010}
& 0.211{\scriptsize$\pm$0.004}
& 0.087{\scriptsize$\pm$0.010}
& 0.161{\scriptsize$\pm$0.015} \\

No reuse
& 0.492{\scriptsize$\pm$0.043}
& \textbf{0.319}{\scriptsize$\pm$0.019}
& 0.390{\scriptsize$\pm$0.062}
& 0.276{\scriptsize$\pm$0.015}
& \textbf{0.522}{\scriptsize$\pm$0.018}
& \textbf{0.256}{\scriptsize$\pm$0.023}
& 0.348{\scriptsize$\pm$0.034}
& 0.267{\scriptsize$\pm$0.014} \\

No KL
& 0.510{\scriptsize$\pm$0.033}
& 0.308{\scriptsize$\pm$0.013}
& 0.320{\scriptsize$\pm$0.077}
& 0.211{\scriptsize$\pm$0.114}
& 0.424{\scriptsize$\pm$0.010}
& 0.243{\scriptsize$\pm$0.005}
& 0.323{\scriptsize$\pm$0.077}
& 0.207{\scriptsize$\pm$0.119} \\

CurriPO
& \textbf{0.512}{\scriptsize$\pm$0.030}
& \textbf{0.319}{\scriptsize$\pm$0.019}
& \textbf{0.415}{\scriptsize$\pm$0.051}
& \textbf{0.280}{\scriptsize$\pm$0.008}
& \textbf{0.522}{\scriptsize$\pm$0.018}
& 0.253{\scriptsize$\pm$0.026}
& \textbf{0.350}{\scriptsize$\pm$0.018}
& \textbf{0.269}{\scriptsize$\pm$0.012} \\
\bottomrule
\end{tabular}
\caption{Ablation on branching and reusing for $N=100$ across three seeds.}
\label{table_8}
\end{table*}

Table~\ref{table_2} shows that this advantage also persists as the population scales to $N{=}100$, with CurriPO again outperforming all baselines under both oracle and noisy feedback. CurriPO also tends to exhibit lower across-seed variability at $N{=}100$, consistent with more stable performance as the population grows. To understand where these gains arise, Figure~\ref{figure_2} visualizes the satisfaction landscape and user distribution for the $N{=}100$ oracle setting (see Appendix~\ref{app:satisfaction-landscape} for the satisfaction landscapes under the remaining population and feedback settings). For the baselines, satisfaction is concentrated near the low-velocity region and fades as the target moves outward; the additional exploration of RM+RND expands this region only slightly. In contrast, CurriPO maintains high satisfaction across a broad portion of the feasible range, showing that its gains arise primarily from users whose target bands lie in the regions that conventional optimization methods tend to underserve. Appendix~\ref{app:welfare} further demonstrates that CurriPO achieves much strong gains under Nash and egalitarian population satisfaction. CurriPO also scales efficiently in terms of training cost: increasing the population from $12$ to $100$ raises its training time only from $1.1$ h to $1.6$ h, whereas the baseline training times increase from $2.6$--$5.7$ h to $22$--$48$ h. Across the eight environments, CurriPO significantly outperforms every baseline under a two-sided Wilcoxon signed-rank test ($p{=}0.0078$), with all comparisons remaining significant after Holm correction ($p_{\mathrm{adj}}{=}0.0391$).

\subsection{Analysis of Branching, Reusing and KL}
We next examine the roles of CurriPO's branching and reusing mechanisms. We compare CurriPO against two ablations: \textbf{Chain}, which removes both branching and reusing and traverses the curriculum along a single path, and \textbf{No reuse}, which retains branching but prevents a reward model from being selected more than once. As shown in Tables~\ref{table_7} and~\ref{table_8}, Chain consistently performs substantially worse than CurriPO, while No reuse closes much of the gap but remains slightly worse in several settings. These results suggest that branching is particularly important for reaching diverse, hard-to-reach users, while reusing provides additional flexibility to recover users that become more accessible from later checkpoints.

To further examine the benefit of reusing, we conduct user-level case studies on how initialization affects reward model optimization. As shown in Figure~\ref{figure_4}, optimizing the same reward model can lead to markedly different outcomes depending on the initialization: in some cases, ground-truth satisfaction decreases from an earlier checkpoint, but increases substantially when the same reward model is reused from a policy learned for another user. This illustrates how reusing allows CurriPO to exploit newly discovered, more favorable initializations for previously difficult users. 

Finally, we ablate the KL penalty to the reference policy in Eq.~\eqref{eq:stage} (\textbf{No KL}), where each stage optimizes from $\pi_{w_k}$ without being constrained toward it. Tables~\ref{table_7} and~\ref{table_8} show that removing it degrades satisfaction in most settings and inflates the variance. We attribute this to the role of the penalty in keeping successive stages as local transitions, which prevents the traversal from abruptly leaving the curriculum it has built.

\subsection{Additional Analysis of Hard-to-Reach Users}
In the previous section, we showed that CurriPO's gains arise primarily in regions that are not reachable by the existing baselines. A natural objection is that the baselines are merely undertrained on these users. Figure~\ref{figure_3} examines this possibility by analyzing the training dynamics on the targets that RM and RM+RND serve most poorly. To this end, we scale the training budget for RM from $0.4$M to $3.2$M steps per user, well beyond the default $1.2$M steps used for both RM and RM+RND. Increasing the training budget, however, leaves these targets essentially unserved: satisfaction at target velocity $5$ remains at $0.001$ throughout, while at target velocity $3.75$, satisfaction does not improve monotonically with additional training. An exploration bonus helps but does not close the gap.

\subsection{Additional Ablations and Sensitivity Analysis}
We conduct additional experiments on checkpoint serving, the traversal order, and $c_{\mathrm{lo}}$, along with a sensitivity analysis of $c_{\mathrm{lo}}$. At serving time, CurriPO assigns each user the wake checkpoint that best matches their reward model. To isolate the effect of this serving rule, we apply the same rule to each baseline over its own policy pool. This improves baselines slightly, but explains only a small fraction of CurriPO's overall gain, indicating that the benefit comes mainly from the traversal that expands the policy pool. We next examine whether CurriPO's traversal order itself is important. For the targets in Figure~\ref{figure_3} that are difficult to reach using conventional optimization, we randomize the curriculum followed by CurriPO. This tests whether warm-starting from other users' policies alone is sufficient, or whether CurriPO's specific traversal order is important. Randomizing the order sharply reduces satisfaction, and the performance gap widens as the target becomes more difficult, demonstrating the importance of CurriPO's curriculum ordering. We further ablate $c_{\mathrm{lo}}$; removing this restriction substantially degrades performance, confirming the importance of restricting expansion to an appropriate frontier. Finally, performance remains stable around the default $c_{\mathrm{lo}}$, suggesting that CurriPO does not require careful tuning of this threshold. Detailed setups and results are provided in Appendix~\ref{app:add}.

\section{Conclusion}
This paper highlights the severity of the problem of unserved users that can arise when optimizing a reward model over a diverse user population, and introduces CurriPO, which addresses this issue by treating the diverse user population as a curriculum. To our knowledge, it is the first to exploit multi-user structure at the optimization stage. By growing a tree-structured traversal with branching and reusing, CurriPO serves an entire population within a single pass. Through extensive experiments on personalized continuous control, we demonstrate that CurriPO achieves higher population satisfaction than baselines at substantially lower training cost, with the gains concentrated on users that conventional optimization struggles to reach. Future work includes extending CurriPO to broader user alignment settings, such as large language models and vision-language-action models, and evaluating it with diverse real-user feedback.

\bibliography{aaai2027}
\clearpage
\appendix
\section*{Appendix}
\setcounter{secnumdepth}{2}

\renewcommand{\thesection}{\Alph{section}}
\renewcommand{\thesubsection}{\thesection.\arabic{subsection}}

\section{Additional Implementation Details}
\label{app:impl}

\subsection{Preference Noise Configuration}
\label{app:bpref}
 
Both the oracle and the noisy pairwise labels judge a segment
$\xi = ((s_t, a_t))_{t=1}^{L}$ through a weighted segment return
\begin{equation}
R_u(\xi) \;=\; \sum_{t=1}^{L} w_t \, u_t,
\qquad
u_t \;=\; g_u(d_t),
\label{eq:teacher-return}
\end{equation}
where $g_u$ is the ground-truth per-step utility of user $u$ defined in Eq.~\eqref{eq:per-step-utility} and $d_t$ is the behavioral feature vector at step $t$.

The oracle pairwise label uses uniform weights $w_t\!=\!1$ and returns the deterministic label
$y = \mathbf{1}\!\left[R_u(\xi^1) > R_u(\xi^0)\right]$.
 
The noisy pairwise label follows the design of B-Pref \citep{lee2021bpref} and applies all five perturbations---myopia, stochasticity, mistakes, skipping, and equal labels---to every query. Writing $\Delta R = R_u(\xi^1) - R_u(\xi^0)$ and letting $\bar{R}_u$ denote the mean segment return of user $u$ over the behavior pool $\mathcal{D}$, each query $(\xi^0, \xi^1)$ is processed in the following order.

\begin{itemize}
\setlength{\itemsep}{1pt}
\item \textbf{Myopic weighting.} The return in Eq.~\eqref{eq:teacher-return} is computed with the
      discounted weights $w_t = \gamma^{\,L-t}$, so that later timesteps within a segment carry
      larger weight and the teacher judges primarily on the terminal behavior of the segment.
\item \textbf{Skip.} If neither segment is informative, i.e.
      $\max\{R_u(\xi^0), R_u(\xi^1)\} < \delta_{\mathrm{skip}} \bar{R}_u$, the query is discarded
      and no label is produced.
\item \textbf{Equal.} If the two segments are nearly indistinguishable, i.e.
      $|\Delta R| < \delta_{\mathrm{equal}} \bar{R}_u$, the query receives the soft label $y = 0.5$,
      which contributes to Eq.~\eqref{eq:bcce} as an evenly split target.
\item \textbf{Stochastic labeling.} Otherwise the label is drawn as
      $y \sim \mathrm{Bernoulli}\big(\sigma(\beta \, \Delta R)\big)$ with rationality constant
      $\beta$, rather than taken deterministically from the sign of $\Delta R$.
\item \textbf{Mistake.} Finally, every hard label is flipped independently with probability
      $p_{\mathrm{flip}}$.
\end{itemize}
 
Table~\ref{tab:bpref} lists the values used for the five components. This
configuration perturbs the preference labels in five qualitatively
different ways at once, and therefore yields reward models whose
geometry deviates substantially from the oracle geometry.
 
\subsection{Baseline Implementations}
\label{app:baselines}
 
\paragraph{Common protocol.}
Every baseline receives exactly the same supervision as CurriPO: RM and RM+RND reuse the identical bank $\{f_u\}_{u \in \mathcal{U}}$, while VPL, CPL, and IPL are fit from the same per-user preference dataset. All methods that interact with the environment share the PPO implementation reported in the main text, start from the same initial policy $\pi_0$, and consume the pool-standardized reward of Eq.~\eqref{eq:zscore}.
 
\paragraph{RM.}
The conventional per-user pipeline \citep{christiano2017deep} runs one PPO job per user on that user's reward model $f_u$ for $1.2$M environment steps. The policy is initialized at $\pi_0$ at the start of each run.
 
\paragraph{RM+RND.}
RM+RND augments RM with a Random Network Distillation bonus \citep{burda2018exploration} of coefficient $1.0$. The target and predictor networks are width-$256$ MLPs producing $64$-dimensional features, and the predictor is trained online on visited states. The per-user budget is again $1.2$M steps. This
baseline isolates whether insufficient exploration alone accounts for the users that RM fails to serve.

\begin{table}[t]
\centering
\footnotesize
\setlength{\tabcolsep}{6pt}
\begin{tabular}{@{}ll@{}}
\toprule
Component & Parameter \\
\midrule
Myopic     & $\gamma = 0.9$ \\
Skip       & $\delta_{\mathrm{skip}} = 0.1$ \\
Equal      & $\delta_{\mathrm{equal}} = 0.1$ \\
Stochastic & $\beta = 12$ \\
Mistake    & $p_{\mathrm{flip}} = 0.1$ \\
\bottomrule
\end{tabular}
\caption{Parameters of the B-Pref noise.}
\label{tab:bpref}
\end{table}
 
\paragraph{VPL.}
VPL \citep{poddar2024personalizing} fits a single latent-conditional reward model over all users jointly. An encoder $q(z \mid \mathcal{A})$ maps an annotation context $\mathcal{A}$ of $8$ preference pairs to a latent $z \in \mathbb{R}^{16}$, and a decoder $r(s,a,z)$ scores state--action pairs conditioned on it. Both are $3$-layer width-$256$ MLPs, trained for $15$k Adam steps with a Bradley--Terry reconstruction loss and an annealed KL term against the latent prior. Each user $u$ is then served by the same $1.2$M-step per-user PPO run used for RM, applied to the conditioned reward $r(\cdot,\cdot,z_u)$.
 
\paragraph{CPL.}
CPL \citep{hejna2024contrastive} first behavior-clones a Gaussian policy with two hidden layers of width $256$ on the pooled dataset $\mathcal{D}$ for $15$k updates. For each user $u$, it then fine-tunes the policy for an additional $45$k updates on the user's preference pairs using the biased contrastive advantage objective, with temperature $\alpha = 0.1$, conservative bias $0.5$, and batch size $96$.

\paragraph{IPL.}
IPL \citep{hejna2023inverse} trains twin $Q$-networks, a $V$-network, and an actor with two hidden layers of width $256$ for $60$k gradient steps on the frozen pooled dataset. The value function is learned using IQL-style expectile regression with expectile $0.7$, while the implicit reward $r = Q - \gamma V$, induced by the inverse Bellman operator, is optimized using the Bradley--Terry preference loss with a $\chi^2$ regularizer. The actor is then extracted via advantage-weighted regression with temperature $\beta = 1/3$.

\begin{table*}[t]
\centering
\small
\setlength{\tabcolsep}{3pt}
\renewcommand{\arraystretch}{0.9}
\setlength{\aboverulesep}{1pt}
\setlength{\belowrulesep}{1pt}
\begin{tabular}{lcccccccc}
\toprule
\multirow{2}{*}{Method} & \multicolumn{4}{c}{Oracle} & \multicolumn{4}{c}{B-Pref noise} \\
\cmidrule(lr){2-5} \cmidrule(lr){6-9}
& HC-1D & HC-2D & H-1D & H-2D
& HC-1D & HC-2D & H-1D & H-2D \\
\midrule
RM
& \dv{0.364}{+0.023}
& \dv{0.238}{+0.023}
& \dv{0.272}{+0.022}
& \dv{0.185}{+0.010}
& \dv{0.328}{+0.017}
& \dv{0.226}{+0.006}
& \dv{0.326}{+0.017}
& \dv{0.200}{+0.037} \\

+RND
& \dv{0.419}{+0.031}
& \dv{0.268}{+0.029}
& \dv{0.275}{+0.015}
& \dv{0.168}{+0.011}
& \dv{0.331}{+0.011}
& \dv{0.172}{+0.004}
& \dv{0.317}{+0.005}
& \dv{0.166}{+0.010} \\

CPL
& \dv{0.201}{+0.013}
& \dv{0.121}{+0.030}
& \dv{0.224}{+0.036}
& \dv{0.102}{+0.026}
& \dv{0.225}{+0.031}
& \dv{0.141}{+0.045}
& \dv{0.212}{+0.032}
& \dv{0.111}{+0.026} \\

IPL
& \dv{0.305}{+0.010}
& \dv{0.209}{+0.016}
& \dv{0.243}{+0.015}
& \dv{0.178}{+0.053}
& \dv{0.323}{+0.019}
& \dv{0.210}{+0.017}
& \dv{0.262}{+0.009}
& \dv{0.139}{+0.011} \\

VPL
& \dv{0.217}{+0.008}
& \dv{0.101}{+0.014}
& \dv{0.257}{-0.009}
& \dv{0.108}{+0.009}
& \dv{0.219}{+0.027}
& \dv{0.130}{+0.016}
& \dv{0.266}{+0.002}
& \dv{0.116}{+0.018} \\

\midrule
CurriPO
& \textbf{0.514}
& \textbf{0.318}
& \textbf{0.355}
& \textbf{0.328}
& \textbf{0.386}
& \textbf{0.344}
& \textbf{0.371}
& \textbf{0.340} \\
\bottomrule
\end{tabular}
\caption{Satisfaction for $N=12$ users under CurriPO-style final serving using policy pools trained by different baseline methods. Each cell reports the served satisfaction, with the gain over the baseline's own serving in parentheses.}
\label{table_3}
\end{table*}

\begin{table*}[t]
\centering
\small
\setlength{\tabcolsep}{3pt}
\renewcommand{\arraystretch}{0.9}
\setlength{\aboverulesep}{1pt}
\setlength{\belowrulesep}{1pt}
\begin{tabular}{lcccccccc}
\toprule
\multirow{2}{*}{Method} & \multicolumn{4}{c}{Oracle} & \multicolumn{4}{c}{B-Pref noise} \\
\cmidrule(lr){2-5} \cmidrule(lr){6-9}
& HC-1D & HC-2D & H-1D & H-2D
& HC-1D & HC-2D & H-1D & H-2D \\
\midrule
RM
& \dv{0.360}{+0.008}
& \dv{0.233}{+0.006}
& \dv{0.336}{+0.027}
& \dv{0.156}{+0.014}
& \dv{0.332}{+0.014}
& \dv{0.197}{+0.014}
& \dv{0.253}{+0.009}
& \dv{0.193}{+0.014} \\

+RND
& \dv{0.360}{+0.016}
& \dv{0.237}{+0.016}
& \dv{0.342}{+0.024}
& \dv{0.153}{+0.008}
& \dv{0.338}{+0.022}
& \dv{0.175}{+0.004}
& \dv{0.275}{+0.008}
& \dv{0.172}{+0.020} \\

CPL
& \dv{0.212}{+0.039}
& \dv{0.102}{+0.013}
& \dv{0.204}{+0.013}
& \dv{0.105}{+0.027}
& \dv{0.198}{+0.019}
& \dv{0.103}{+0.010}
& \dv{0.206}{+0.012}
& \dv{0.099}{+0.012} \\

IPL
& \dv{0.311}{+0.012}
& \dv{0.219}{+0.024}
& \dv{0.248}{+0.008}
& \dv{0.143}{+0.015}
& \dv{0.312}{+0.016}
& \dv{0.222}{+0.031}
& \dv{0.275}{+0.014}
& \dv{0.135}{+0.006} \\

VPL
& \dv{0.255}{+0.047}
& \dv{0.116}{+0.009}
& \dv{0.292}{+0.017}
& \dv{0.112}{+0.018}
& \dv{0.222}{+0.007}
& \dv{0.132}{+0.007}
& \dv{0.281}{+0.006}
& \dv{0.107}{+0.014} \\

\midrule
CurriPO
& \textbf{0.512}
& \textbf{0.319}
& \textbf{0.415}
& \textbf{0.280}
& \textbf{0.522}
& \textbf{0.253}
& \textbf{0.350}
& \textbf{0.269} \\
\bottomrule
\end{tabular}
\caption{Satisfaction for $N=100$ users under CurriPO-style final serving using policy pools trained by different baseline methods. Each cell reports the served satisfaction, with the gain over the baseline's own serving in parentheses.}
\label{table_4}
\end{table*}

\section{Additional Analysis on Serving from the Wake and Sensitivity Analysis}
\label{app:add}

\begin{table}[t]
\centering\small
\begin{tabular}{lccccc}
\toprule
\multirow{2}{*}{Method} & \multicolumn{5}{c}{Target velocity} \\
\cmidrule(lr){2-6}
& 2 & 3.75 & 4.5 & 4.75 & 5 \\
\midrule

Random Order
& 0.428
& 0.212
& 0.159
& 0.147
& 0.128 \\
CurriPO
& \textbf{0.613}
& \textbf{0.559}
& \textbf{0.433}
& \textbf{0.441}
& \textbf{0.445} \\
\bottomrule
\end{tabular}
\caption{Random-ordering ablation on hard-to-reach targets.}
\label{tab:model-version-comparison}
\end{table}

\subsection{Analysis on Serving from the Wake}
CurriPO serves each user the wake checkpoint their own reward model prefers, which raises the question of how much of its advantage comes from selecting over a pool rather than from the traversal that built the pool. Tables~\ref{table_3} and~\ref{table_4} give every baseline the same serving rule over its own policy pool. Nearly all baseline-setting pairs improve, confirming that cross-user serving is useful on its own, but the gains are small and leave a wide margin. The reason is that selection and traversal play different roles. Selection is a repair mechanism: when a user's own optimization fails, it lets that user fall back on a neighbor's policy that happens to score better under their reward model, recovering part of the loss within the behaviors the policy pool already contains. It cannot, however, extend the pool. Under per-user optimization, every policy in it was produced from $\pi_0$, so regions unreachable in that way remain unrepresented no matter how the serving rule redistributes them, and the users who lie there gain nothing. CurriPO's traversal instead changes what is available to select from, because each stage starts from a new checkpoint that enables higher-quality exploration of previously unexplored regions.

\subsection{Analysis on Ordering}
CurriPO reaches a given user through a specific sequence of intermediate
checkpoints. To test whether this ordering itself matters, we construct a \textbf{Random Order} baseline that takes the curriculum CurriPO traversed to reach each of the hard targets in Figure~\ref{figure_3} and randomly permutes the order in which its intermediate reward models are optimized, running the same number of KL-anchored local stages in Eq.~(6) with the same per-stage PPO budget and warm-starting each stage from the preceding checkpoint. Table~\ref{tab:model-version-comparison} reports utilitarian satisfaction on $N{=}12$ under oracle feedback. Random Order is uniformly worse. Warm-starting from other users' policies is thus not sufficient on its own---the progression discovered by CurriPO's coverage-based selection is what makes hard-to-optimize users reachable.

\subsection{Analysis on $c_{\mathrm{lo}}$}
Table~\ref{tab:clo_sweep_n100_util} varies and ablate the key parameter $c_{\mathrm{lo}}$. Setting $c_{\mathrm{lo}}{=}0$ admits every user as a candidate at every checkpoint, causing Eq.~\eqref{eq:pick} to reduce to always targeting the least-covered user. As a result, mean satisfaction drops from $0.365$ to $0.208$, demonstrating that removing the threshold substantially degrades performance. These results suggest that restricting expansion to an appropriate curriculum frontier is important for effective exploration. At the same time, performance is relatively insensitive around the default setting of $c_{\mathrm{lo}}{=}0.60$: across $c_{\mathrm{lo}}\in[0.50,0.70]$, mean satisfaction varies only from $0.356$ to $0.365$, indicating that CurriPO does not require careful tuning of $c_{\mathrm{lo}}$.

\section{Additional Experimental Details}
\subsection{Additional User Satisfaction Visualization}
\label{app:satisfaction-landscape}
\begin{figure*}[t]
    \centering
    \includegraphics[width=\linewidth]{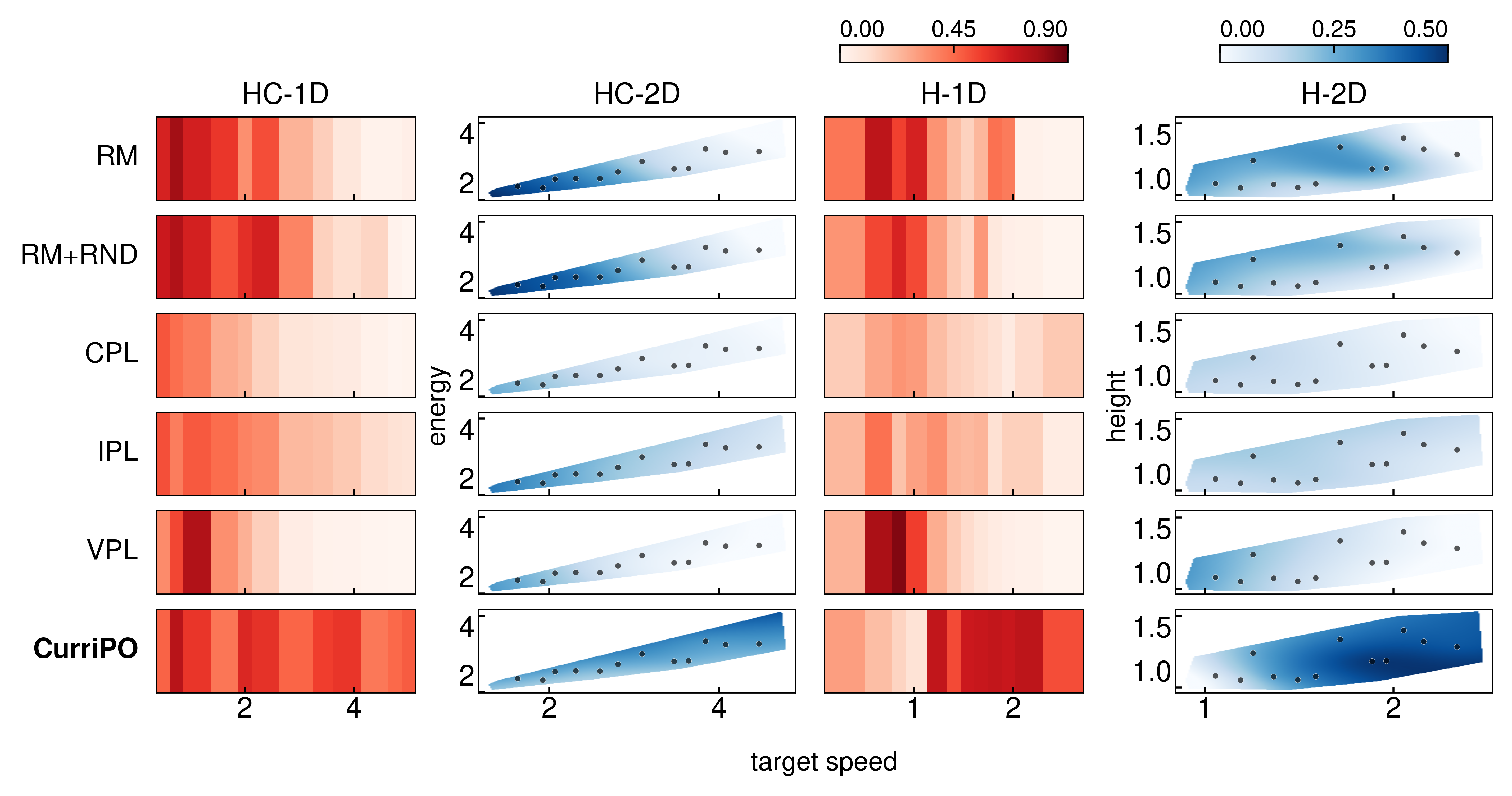}
    \caption{\textbf{User satisfaction visualization across different optimization methods with $N=12$ users under oracle feedback.} User satisfaction landscapes are shown for HalfCheetah and Hopper under 1-D and 2-D preference settings, with target speed as the shared preference dimension and energy or height as the second dimension. Darker colors indicate higher user satisfaction.}

    \label{figure_2_2}
\end{figure*}

\begin{figure*}[t]
    \centering
    \includegraphics[width=\linewidth]{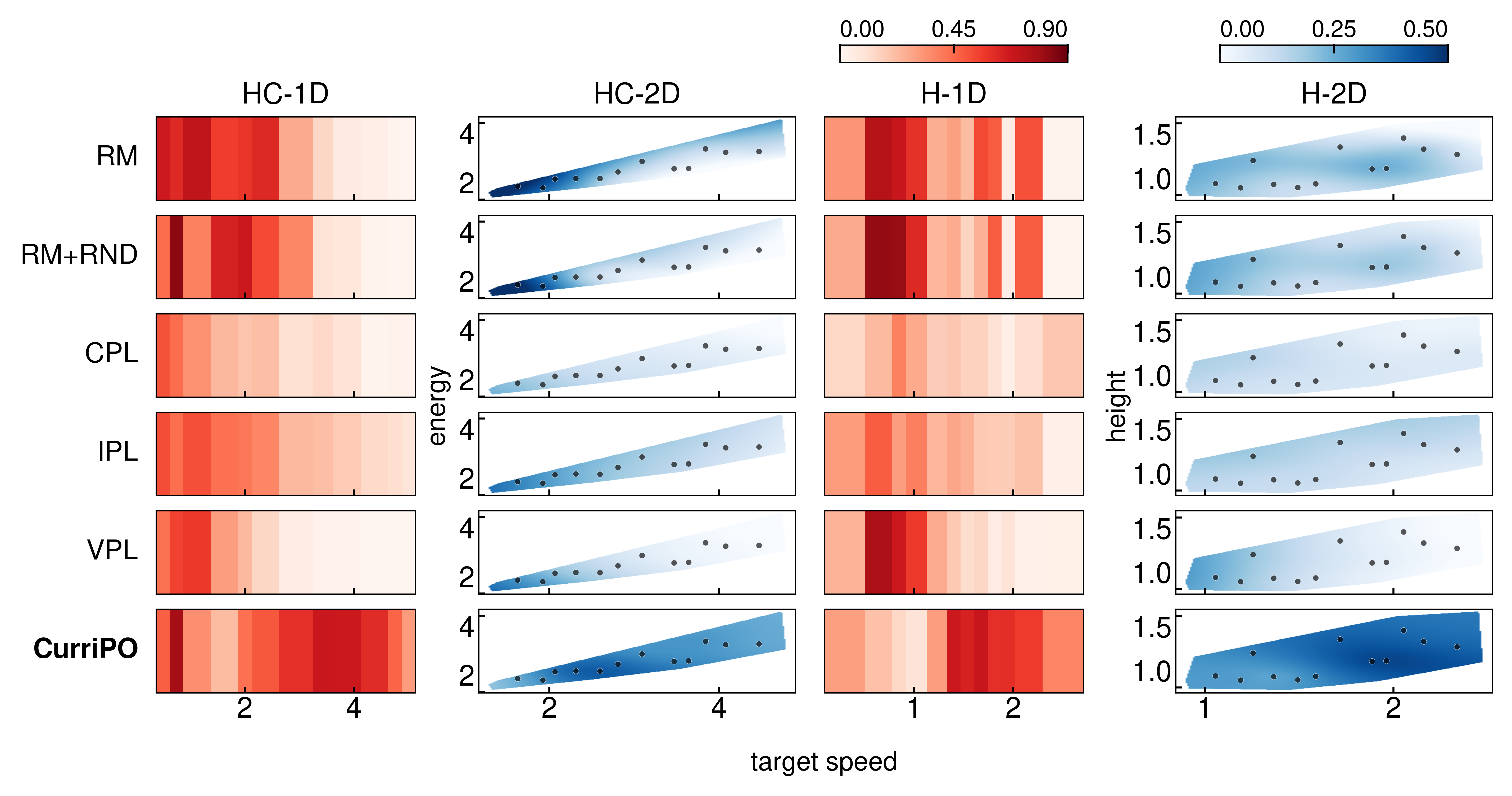}
    \caption{\textbf{User satisfaction visualization across different optimization methods with $N=12$ users under noisy feedback.} User satisfaction landscapes are shown for HalfCheetah and Hopper under 1D and 2D preference settings, with target speed as the shared preference dimension and energy or height as the second dimension. Darker colors indicate higher user satisfaction.}
    \label{figure_2_3}
\end{figure*}

\begin{figure*}[t]
    \centering
    \includegraphics[width=\linewidth]{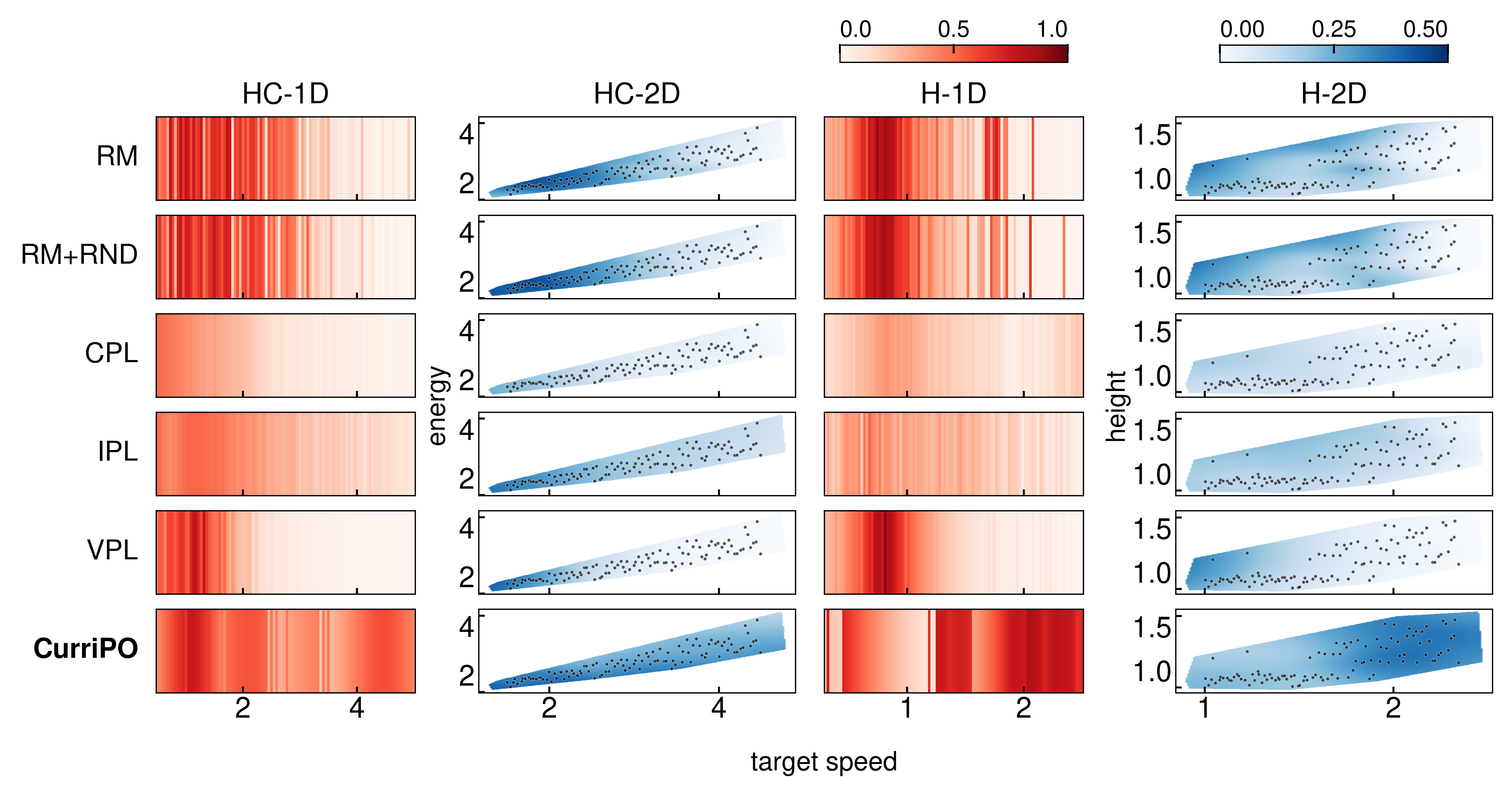}
    \caption{\textbf{User satisfaction visualization across different optimization methods with $N=100$ users under noisy feedback.} User satisfaction landscapes are shown for HalfCheetah and Hopper under 1D and 2D preference settings, with target speed as the shared preference dimension and energy or height as the second dimension. Darker colors indicate higher user satisfaction.}
    \label{figure_2_4}
\end{figure*}

Figure~\ref{figure_2} in the main text reports the user satisfaction landscape for the $N{=}100$ population under oracle feedback. Here we provide the same visualization for the remaining three
population--feedback configurations: $N{=}12$ under oracle feedback (Figure~\ref{figure_2_2}),
$N{=}12$ under B-Pref noisy feedback (Figure~\ref{figure_2_3}), and $N{=}100$ under B-Pref noisy feedback (Figure~\ref{figure_2_4}).

\subsection{Nash and Egalitarian Population Satisfaction}
\label{app:welfare}

\begin{table*}[t]
\centering
\footnotesize
\setlength{\tabcolsep}{2pt}
\renewcommand{\arraystretch}{0.9}
\setlength{\abovecaptionskip}{4pt}
\setlength{\belowcaptionskip}{0pt}
\begin{tabular}{lcccccccccccccccc}
\toprule
 & \multicolumn{4}{c}{Oracle ($N{=}12$)} & \multicolumn{4}{c}{Noisy ($N{=}12$)}
 & \multicolumn{4}{c}{Oracle ($N{=}100$)} & \multicolumn{4}{c}{Noisy ($N{=}100$)} \\
\cmidrule(lr){2-5}\cmidrule(lr){6-9}\cmidrule(lr){10-13}\cmidrule(lr){14-17}
Method
 & HC-1D & HC-2D & H-1D & H-2D
 & HC-1D & HC-2D & H-1D & H-2D
 & HC-1D & HC-2D & H-1D & H-2D
 & HC-1D & HC-2D & H-1D & H-2D \\
\midrule
\multicolumn{17}{l}{\textit{Egalitarian}} \\
RM
 & 0.018 & 0.010 & 0.012 & 0.014 & 0.005 & 0.002 & 0.013 & 0.009
 & 0.001 & 0.018 & 0.008 & 0.011 & 0.000 & 0.001 & 0.005 & 0.006 \\
RM+RND
 & 0.002 & 0.018 & 0.008 & 0.016 & 0.004 & 0.002 & 0.010 & 0.020
 & 0.001 & 0.013 & 0.012 & 0.011 & 0.000 & 0.001 & 0.010 & 0.008 \\
CPL
 & 0.006 & 0.013 & 0.050 & 0.014 & 0.006 & 0.022 & 0.034 & 0.027
 & 0.006 & 0.004 & 0.012 & 0.007 & 0.003 & 0.002 & 0.001 & 0.006 \\
IPL
 & 0.089 & 0.088 & 0.043 & \textbf{0.062} & 0.075 & 0.085 & 0.035 & 0.060
 & 0.075 & 0.074 & 0.014 & 0.019 & 0.048 & 0.057 & 0.011 & 0.017 \\
VPL
 & 0.002 & 0.005 & 0.008 & 0.012 & 0.001 & 0.007 & 0.013 & 0.011
 & 0.002 & 0.003 & 0.008 & 0.009 & 0.002 & 0.002 & 0.008 & 0.009 \\
CurriPO
 & \textbf{0.348} & \textbf{0.212} & \textbf{0.102} & 0.037
 & \textbf{0.139} & \textbf{0.201} & \textbf{0.075} & \textbf{0.091}
 & \textbf{0.235} & \textbf{0.155} & \textbf{0.055} & \textbf{0.077}
 & \textbf{0.206} & \textbf{0.119} & \textbf{0.052} & \textbf{0.058} \\
\midrule
\multicolumn{17}{l}{\textit{Nash}} \\
RM
 & 0.183 & 0.128 & 0.222 & 0.118 & 0.162 & 0.107 & 0.266 & 0.105
 & 0.196 & 0.162 & 0.186 & 0.102 & 0.132 & 0.086 & 0.212 & 0.141 \\
RM+RND
 & 0.192 & 0.158 & 0.146 & 0.115 & 0.140 & 0.067 & 0.235 & 0.125
 & 0.175 & 0.149 & 0.187 & 0.103 & 0.107 & 0.080 & 0.230 & 0.118 \\
CPL
 & 0.087 & 0.064 & 0.158 & 0.059 & 0.108 & 0.080 & 0.146 & 0.075
 & 0.098 & 0.064 & 0.151 & 0.063 & 0.113 & 0.072 & 0.153 & 0.073 \\
IPL
 & 0.251 & 0.178 & 0.182 & 0.118 & 0.257 & 0.176 & 0.206 & 0.119
 & 0.264 & 0.182 & 0.194 & 0.118 & 0.254 & 0.174 & 0.213 & 0.117 \\
VPL
 & 0.051 & 0.042 & 0.126 & 0.064 & 0.043 & 0.057 & 0.142 & 0.064
 & 0.056 & 0.049 & 0.136 & 0.062 & 0.064 & 0.061 & 0.132 & 0.062 \\
CurriPO
 & \textbf{0.505} & \textbf{0.310} & \textbf{0.302} & \textbf{0.249}
 & \textbf{0.342} & \textbf{0.335} & \textbf{0.309} & \textbf{0.307}
 & \textbf{0.500} & \textbf{0.311} & \textbf{0.346} & \textbf{0.241}
 & \textbf{0.503} & \textbf{0.244} & \textbf{0.295} & \textbf{0.232} \\
\bottomrule
\end{tabular}
\caption{Egalitarian and Nash population satisfaction on $N{=}12$ and $N{=}100$ over 3 seeds.}
\label{tab:welfare}
\end{table*}

\begin{table}[t]
\centering
\setlength{\tabcolsep}{4pt}
\renewcommand{\arraystretch}{0.9}
\setlength{\aboverulesep}{1pt}
\setlength{\belowrulesep}{1pt}
\setlength{\abovecaptionskip}{4pt}
\setlength{\belowcaptionskip}{0pt}
{\scriptsize
\begin{tabular}{lcccccccc}
\toprule
 & \multicolumn{4}{c}{\footnotesize Oracle ($N=100$)} & \multicolumn{4}{c}{\footnotesize Noisy ($N=100$)} \\
\cmidrule(lr){2-5}\cmidrule(lr){6-9}
$c_{\mathrm{lo}}$
 & HC-1D & HC-2D & H-1D & H-2D
 & HC-1D & HC-2D & H-1D & H-2D \\
\midrule
0.00
 & 0.311 & 0.140 & 0.221 & 0.087
 & 0.378 & 0.234 & 0.199 & 0.093 \\
0.50
 & 0.493 & 0.174 & 0.474 & 0.292
 & 0.441 & 0.192 & 0.473 & 0.305 \\
0.55
 & 0.445 & 0.327 & 0.525 & 0.286
 & 0.406 & 0.246 & 0.412 & 0.206 \\
0.60
 & 0.512 & 0.319 & 0.415 & 0.280
 & 0.522 & 0.253 & 0.350 & 0.269 \\
0.65
 & 0.428 & 0.332 & 0.459 & 0.206
 & 0.477 & 0.253 & 0.445 & 0.285 \\
0.70
 & 0.452 & 0.326 & 0.463 & 0.261
 & 0.487 & 0.226 & 0.337 & 0.292 \\
\bottomrule
\end{tabular}
}
\caption{Sensitivity and ablation analysis of $c_{\mathrm{lo}}$ on $N{=}100$.}
\label{tab:clo_sweep_n100_util}
\end{table}

\paragraph{Metrics.}
The utilitarian satisfaction reported in the main text,
$W_{\mathrm{util}} = \frac{1}{N}\sum_{u} G_u({\pi_u})$, weighs
every user equally but is insensitive to how satisfaction is
\emph{distributed}: a method that serves part of the population well
and abandons the rest can attain the same mean as a method that serves
everyone moderately. Since the concern motivating this work is
precisely that some users are left unserved, we additionally report two
distribution-sensitive criteria,
\begin{equation}
W_{\mathrm{nash}} = \Big(\prod_{u=1}^{N} G_u(\pi_u)\Big)^{1/N},
\qquad
W_{\mathrm{egal}} = \min_{u} G_u(\pi_u).
\label{eq:welfare}
\end{equation}
The egalitarian criterion is determined by the single worst-served user. The Nash
criterion sits between the egalitarian and utilitarian: being multiplicative, it is driven toward zero by any user receiving near-zero satisfaction, while still
rewarding broad gains. Tables~\ref{tab:welfare} reports both criteria for $N{=}12$ and $N{=}100$ under oracle and B-Pref noisy feedback. CurriPO achieves the highest Nash welfare in all 16 settings and the highest egalitarian welfare in 15 of 16. The advantage is particularly pronounced under the egalitarian criterion, where most baselines leave at least one user with near-zero satisfaction, whereas CurriPO substantially raises the satisfaction of the worst-served users. These results indicate that CurriPO's gains are not driven by concentrating improvements on already well-served users, but by extending effective coverage to users who would otherwise remain underserved.

\end{document}